# Integrating mobile and fixed monitoring data for high-resolution $PM_{2.5}$ mapping using machine learning

Rui Xu[1,2,3], Dawen Yao[1,2,3], Yuzhuang Pian[1,2,3], Ruhui Cao[1,2,3], Yixin Fu[1,2,3], Xinru Yang[1,2,3], Ting Gan[1,2,3], Yonghong Liu*[1,2,3]

1. School of Intelligent Systems Engineering, Sun Yat-sen University, Shenzhen, 518107, Guangdong, China.
2. Guangdong Provincial Key Laboratory of Intelligent Transportation System, Sun Yat-Sen University, Guangzhou 510275, China.
3. Guangdong Provincial Engineering Research Center for Traffic Environmental Monitoring and Control, Guangzhou, 510275, Guangdong, China.

Abstract: Constructing high resolution air pollution maps at lower cost is crucial for sustainable city management and public health risk assessment. However, traditional fixed-site monitoring lacks spatial coverage, while mobile low-cost sensors exhibit significant data instability. This study integrates $PM_{2.5}$ data from 320 taxi-mounted mobile low-cost sensors and 52 fixed monitoring stations to address these limitations. By employing the machine learning methods, an appropriate mapping relationship was established between fixed and mobile monitoring concentration. The resulting pollution maps achieved 500-meter spatial and 5-minute temporal resolutions, showing close alignment with fixed monitoring data (+4.35% bias) but significant deviation from raw mobile data (-31.77%). The fused map exhibits the fine-scale spatial variability also observed in the mobile pollution map, while showing the stable temporal variability closer to that of the fixed pollution map (fixed: 1.12±0.73%, mobile: 3.15±2.44%, mapped: 1.01±0.65%). These findings demonstrate the potential of large-scale mobile low-cost sensor networks for high-resolution air quality mapping, supporting targeted urban environmental governance and health risk mitigation.

# 1. Introduction

Air quality is fundamental to urban sustainable development and public health. The sustainable development of cities is closely related to air quality [1]. Meanwhile, the prevalence of severe air pollution in urban areas poses significant risks to population health. The World Health Organization (WHO) reported that 99% of the population lived in an environment exceeding its ambient air quality guideline in 2019 [2]. Air pollution maps with high spatial and temporal resolution will be beneficial for fine-grained air pollution management. Also, populations need more refined pollution map for exposure risk assessment.

Comprehensive cognition of urban pollution is an effective prerequisite for the improvement of urban air quality. The air pollution map directly represent the spatial and temporal variability [3,4]. Recent researches have increasingly focused on the development of high-temporal and spatial air pollution maps for urban areas [5]. The general methods for assessing air quality of whole city areas are numerical simulation, remote sensing, and air quality monitoring. Numerical simulation is limited in accurately replicating real-world pollution dispersion patterns and requires substantial computational resources to simulate city-scale areas effectively [6,7]. Remote sensing satellites data is able to estimate $PM_{2.5}$ (Particles less than 2.5μm in diameter) concentration, but suffers from coarse temporal-spatial resolution and data missing [8,9]. The actual pollution concentration can also be obtained by monitoring. Nonetheless, urban monitoring stations typically provide data with limited temporal and spatial coverage due to cost constraints [10,11]. Fixed monitoring system often covers only small-scale areas and fail to provide adequate spatial data. For instance, fixed-route mobile [12–14] and fixed station [15–17] monitoring suffer from insufficient and sparse spatial coverage. Similarly, non-fixed routes of individual vehicles are often temporally sparse and spatially limited [18,19].

The limitations of the aforementioned methods hinder a comprehensive understanding of urban air pollution. However, with the development and cost reduction of low-cost sensor (LCS) [20], it is possible to construct the high resolution pollution map through large-scale mobile monitoring network. Vehicle-mounted acquisition has emerged as a robust mobile monitoring method, capable of effectively continuous tracking of pollutant concentrations over extended periods and across extensive spatial areas [21]. Dense observation in both temporal and spatial dimensions are essential for constructing high resolution pollution maps. Several studies have explored the use of mobile monitoring to enhance awareness of pollution levels and develop detailed pollution maps in urban areas. For instance, a high resolution 2D gridded air pollution map was developed using mobile monitoring in a small, densely populated urban area to aid exposure assessment studies [22]. Similarly, data from mobile LCS monitoring with PMS7003 were integrated with other sources to map the grid distribution of $PM_{2.5}$ in a micro-scale walkable environment [23], helping to identify hotspots of air pollution and suggest healthier walking routes. A machine learning model, integrating $NO_2$ measurement data from air quality monitoring stations (AQMS), fixed LCS, and the satellite retrievals alongside environmental covariates, was employed to map hourly ground-level $NO_2$ concentrations at a 1 km resolution, facilitating the identification of pollution hotspots[24]. Furthermore, it was demonstrated the effectiveness of mobile monitoring with LCS in tracking changes in urban air quality[25–29]. These findings underscore the value of mobile monitoring for source attribution, traceability, and potential mitigation strategies at the urban micro-scale. The integration of mobile monitoring data with other sources has proven valuable in constructing pollution map and assessing air quality. However, large-scale mobile monitoring data still lacks application and exploration at the citywide level. The characteristics and disparities of citywide high resolution pollution maps, based on large-scale mobile monitoring data, require further investigation. This research is essential to tackle challenges related to modeling accuracy and to investigate innovative approaches for pollution mapping utilizing large-scale LCS mobile monitoring technologies.

    Therefore, this study aims to develop the high resolution $PM_{2.5}$ pollution map by integrating large-scale mobile monitoring data with fixed monitoring and urban characterization data. Enhanced accuracy in air pollution mapping is anticipated,

offering reliable data to support individual health exposure assessments and environmental governance. Initially, the reliability of mobile monitoring results was assessed by comparison with fixed monitoring data. Then, by analyzing the correlation between mobile and fixed monitoring data under different spatial and temporal conditions, the optimal resolution for pollution map construction was determined. Considering the actual scenario of the city, three different pollution maps were constructed from the monitoring data, and their spatial and temporal characteristics were systematically analyzed.

# 2. Data and methods

## 2.1. Study area

Guangzhou city, located in Guangdong-Hongkong-Macau Great Bay Area (GBA), China, is a megacity with resident population of over 18.83 million. Six central districts of Guangzhou city, had been focused including Liwan, Yuexiu, Haizhu, Tianhe, Baiyun, and Huangpu (Fig.1). With a total area of 1559.59 km², this region comprises only 20.98% of Guangzhou's land area. However, it is home to approximately 11.13 million residents, making up 59.11% of the city's total population and exhibiting a significant population density of 7136 inhabitants per km². Over the past three years, the annual average $PM_{2.5}$ concentration has consistently remained below 25μg/m³. However, localized $PM_{2.5}$ levels along roadways are significantly elevated due to the combined effects of high-density traffic emissions and construction-related activities in proximity to roadside areas [30].

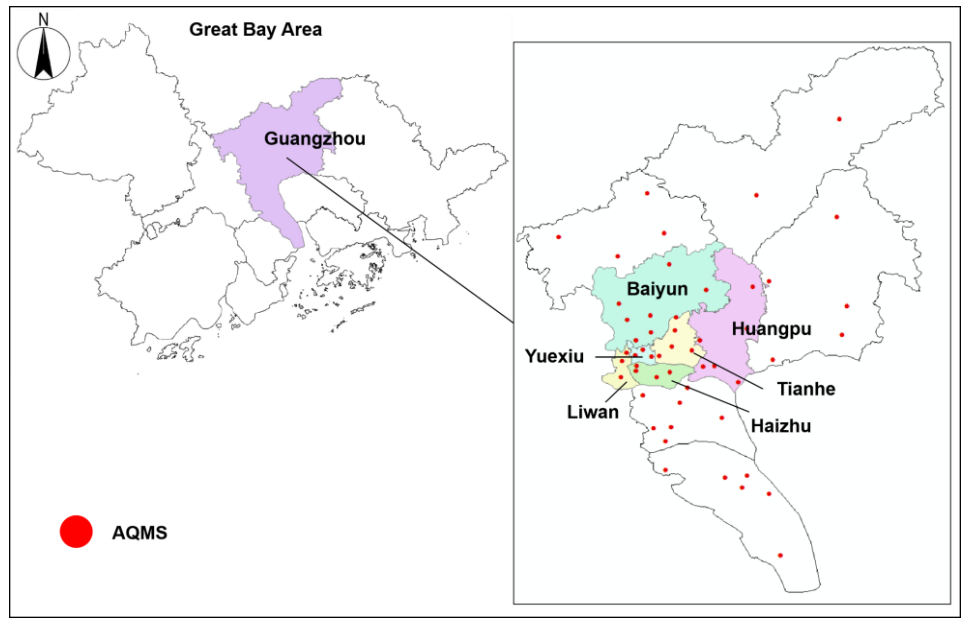

Fig.1 Spatial distribution of AQMS in study area.

## 2.2. Air quality monitoring data

**Fixed AQMS monitoring**: This study utilized fixed monitoring data of $PM_{2.5}$ from 52 AQMSs. Totally 0.46 million valid data of $PM_{2.5}$ concentration were reserved at a 5-minute resolution between March 1 and March 31, 2023. There are 28 stations located within the study area and 24 additional stations located outside this region (Fig.1).

**Mobile LCS monitoring**：LCSs were deployed on the roof lights of 320 electric taxis to obtain mobile monitoring data. Utilizing the taxis' power supply, the system enables continuous 24-hour monitoring, facilitating higher-frequency and longer-duration data collection than earlier self-powered monitoring devices [31]. The sensor's model is PMS5003T[32] with collecting particulate matter at 15-sencond intervals. It uses the laser scattering principle to achieve accurate measurements. This LCS can operate under conditions ranging from -10℃~60℃ and 0~95% relative humidity and measured $PM_{2.5}$ concentrations within a range of 0~500μg/m$^3$ alongside temporal (time) and spatial (latitude, longitude) data [33]. Ambient temperature and humidity were also collected synchronously. A total of 35.12 million mobile monitoring records were retained, representing a volume of data 76 times greater than that obtained from fixed monitoring stations. The spatial distribution of mobile data is presented in Fig.2. The samplings of mobile monitoring are not uniform in space, and the maximum sampling times of grid can reach 8701 samplings. In the average spatiotemporal of the mobile monitoring results, it was observed that concentration data were collected for 20.6% of the grids within an hourly interval, while 5.3% of the grids recorded data within a 5-minute interval. Overall, the experiment achieved a coverage of 77.8% of the grids, with concentration data collected across these areas.

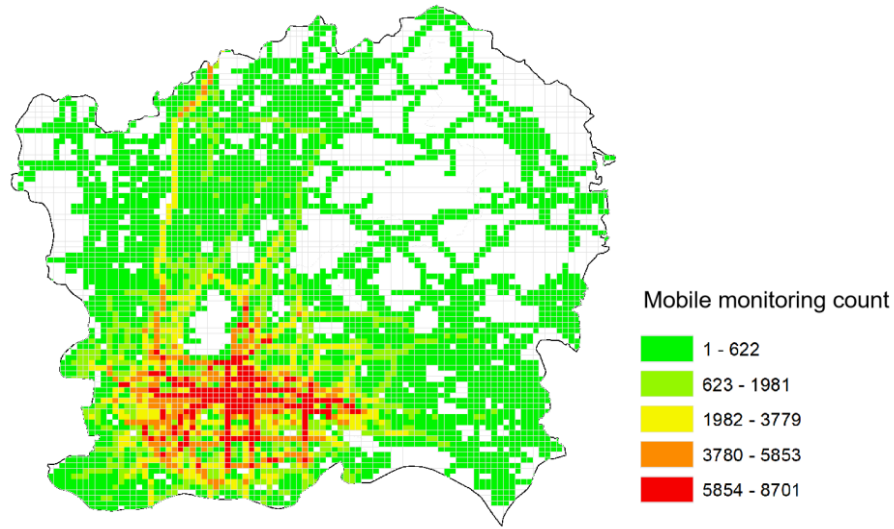

Fig.2 Spatial coverage of mobile monitoring data between March 1 and March 31, 2023.

## 2.3. Calibration for LCS monitoring

Calibration is a critical prerequisite for ensuring the accuracy and reliability of long-term monitoring efforts. To validate the performance of mobile LCS monitoring, data quality control measures were implemented, including the removal of unreasonable concentration values, such as those below 0μg/m³ or exceeding 500μg/m³, which are considered outside the valid measurement range. Besides, a validation test was conducted using a AQMS in Yangji, where three LCS devices collected concurrent concentration data during March, 2023. The concentrations measured by the fixed AQMS served as reference values for error correction in LCS monitoring data. Four widely calibration methods were evaluated to assess their effectiveness in improving data accuracy. First, a linear correction[34,35] was applied to address the systematic bias of LCSs, as shown in Eq(1). Subsequently, calibration method was developed by applying a polynomial linear correction[36] that incorporates the concurrently acquired temperature and humidity data, as shown in Eq(2) and Eq(3).

$$PM_{2.5\_AQMS} = a \cdot PM_{2.5\_LCS} + b \tag{1}$$

$$PM_{2.5\_AQMS} = j_1 \cdot PM_{2.5\_LCS} + j_2 \cdot RH + j_3 \tag{2}$$

$$PM_{2.5\_AQMS} = k_1 \cdot PM_{2.5\_LCS} + k_2 \cdot RH + k_3 \cdot T + k_4 \tag{3}$$

$$PM_{2.5\_AQMS} = xgb\_model([PM_{2.5\_LCS}, RH, T]) \tag{4}$$

Finally, a black-box machine learning model[37], XGBoost, was employed to fit the calibration model, as shown in Eq(4). In Eq(1)-(4), the parameters $a, b, j_1, j_2, j_3, k_1, k_2, k_3, k_4$, and XGBoost model were derived by fitting the LCS concentration data to the corresponding concentrations measured at the Yangji AMQS. In the above equations, $PM_{2.5\_LCS}$ represents the PM$_{2.5}$ concentration measured by the LCS, $PM_{2.5\_AQMS}$ denotes the PM$_{2.5}$ concentration measured by the AQMS, RH stands for relative humidity, and T represents temperature.

80% of the monitoring data was used to fit the model, and the remaining 20% was reserved exclusively for the final evaluation stage, ensuring the rigor of the results. To evaluate the effectiveness of the calibration models, four statistical metrics were employed, including the Pearson correlation coefficient (r), coefficient of determination ($R^2$), mean absolute error (MAE), and root mean square error (RMSE). These metrics are defined sequentially in Eq(5) to Eq(8). $y$ represents the concentration recorded by the Yangji AQMS, whereas $\hat{y}$ denotes the calibrated concentration.

$$r = \frac{\sum_{i=1}^{n}(y_i - \bar{y})(\hat{y}_i - \bar{\hat{y}}_i)}{\sqrt{\sum_{i=1}^{n}(y_i - \bar{y})^2}\sqrt{\sum_{i=1}^{n}(\hat{y}_i - \bar{\hat{y}})^2}} \quad (5)$$

$$R^2 = 1 - \frac{\sum_{i=1}^{n}(y_i - \hat{y}_i)^2}{\sum_{i=1}^{n}(y_i - \bar{y})^2} \quad (6)$$

$$MAE = \frac{\sum_{i=1}^{N}|y_i - \hat{y}_i|}{N} \quad (7)$$

$$RMSE = \sqrt{\frac{1}{n}\sum_{i=1}^{n}(y_i - \hat{y}_i)^2} \quad (8)$$

## 2.4. Data Integration for PM$_{2.5}$ mapping

### 2.4.1. Urban feature data

Urban features served as auxiliary variables that help elucidate the relationship between mobile and fixed monitoring concentrations within each grid. These features, such as land cover, land use, road network, and building area, provide essential contextual information that influences pollution dispersion. Fig. 3 shows four different types of urban feature data. Land cover data was obtained from [38], providing different types of land cover at a spatial resolution of 30-meter. The study area encompasses five different cover categories, including cultivated, forest, grass, water and artificial. Land use data employed in this study was derived from "EULUC-China", as developed by [39]. It contains five major categories including residential, commercial, industrial, transportation and public management and service. Road network data, specifically road lengths were sourced from OpenStreetMap (OSM) to quantify the traffic intensity [40]. Additionally, building area data was incorporated to account for its influence on the dispersion and diffusion of urban pollutants. Urban feature data were systematically quantified within each grid and utilized as input variables in the machine learning model to explore the correlation between mobile monitoring and fixed monitoring concentrations.

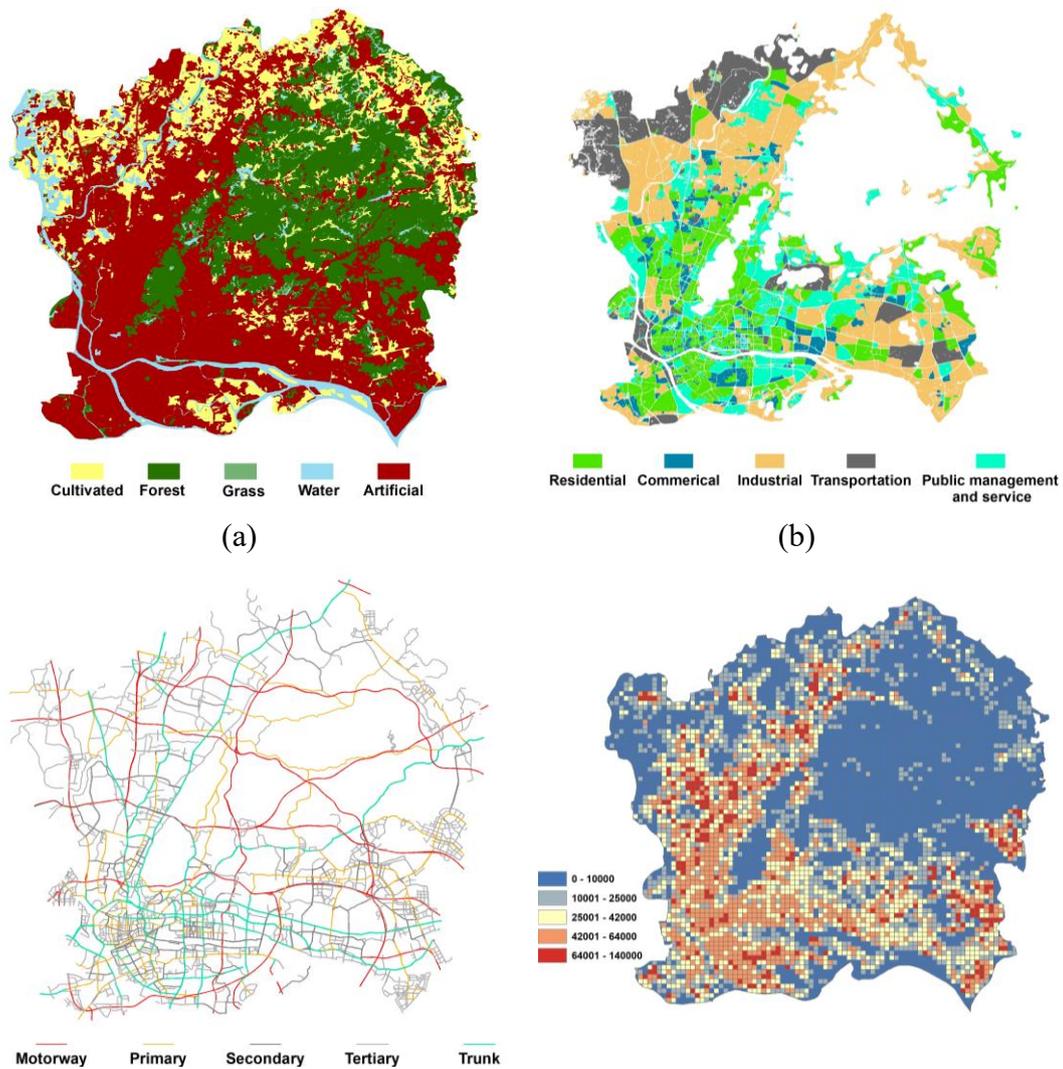

(c) (d)

Fig. 3 Urban feature data: (a) Land cover, (b) Land use, (c) Road network and (d) Building area (unit: m$^2$).

## 2.4.2. Alignment between mobile and fixed monitoring concentrations

Concentrations from mobile and fixed monitoring are spatially distributed differently and temporally collected at different frequencies, and the choice of appropriate spatial and temporal resolution alignment is essential. A fixed AQMS was defined as the geometric center to form a square of specified length. The square was defined as a grid and relation between all mobile and fixed concentrations within the grid was analyzed. The r between fixed and mobile monitoring data was calculated for various spatial and temporal resolutions, including spatial scales of 500, 1000, and 2000 meters, and temporal intervals of 5, 10, 30, and 60 minutes. This analysis aimed to identify the optimal spatial-temporal resolution for constructing the air pollution map. The correlation analysis here gave a reasonable spatial grid size and time interval.

Alignment between concentrations was performed after the spatial-temporal resolution was determined. Based on monitoring data availability(Fig.4b), the grids were grouped into three categories: (a) those containing both mobile and fixed monitoring data, (b) those with only mobile monitoring data, and (c) no available monitoring data. To enhance the spatial heterogeneity and coverage of pollution maps, additional mobile monitoring data were utilized. A range of machine learning models were employed to predict PM$_{2.5}$ concentrations, providing a representation of pollution levels across the grid areas. These predicated concentrations were referred as the mapped concentration. Fig.4(c) presents the structural framework of the PM$_{2.5}$ concentration prediction model. Machine learning was introduced here because it is good at capturing complex non-linear relationships between air quality data [41]. The results of different machine learning models including XGBoost, Random Forest, Lasso, etc. were called mapping models. The mapping models were trained using data from grids to describe the relationship from mobile to the fixed monitoring concentration within the grid.

### 2.4.3. PM$_{2.5}$ mapping

The training dataset for the mapping model was constructed using 52 grids that contained both mobile and fixed monitoring concentration data. As primary feature inputs, key statistical metrics, including the mean, minimum, and maximum values of the mobile monitoring data, were calculated to characterize pollutant distribution within each grid. To enhance the predictive capacity of the model, supplementary urban features, such as building area, diverse land use and land cover classifications, and road length, were integrated to account for their potential influence on the relationship between mobile and fixed monitoring concentrations. As a result, the training dataset incorporated both statistical summaries of mobile monitoring data and urban morphological characteristics. The fixed monitoring concentration were then used as target variables for model training. The r, MAE and mean absolute percentage error (MAPE) are the metrics to assure consistent and biased errors and defined by Eq(9), (7) and (10), respectively. The optimal mapping model was applied to all mobile monitoring concentrations to predict the grids' mapped concentration.

$$r = \frac{\sum_{i=1}^{N}(y_i - \overline{y})(y_i - \overline{\hat{y}})}{\sqrt{\sum_{i=1}^{N}(y_i - \overline{y})^2}\sqrt{\sum_{i=1}^{N}(y_i - \overline{\hat{y}})^2}} \qquad (9)$$

$$MAPE = \frac{1}{N}\sum_{i=1}^{N}\left|\frac{y_i - \hat{y}_i}{y_i}\right| \times 100\% \qquad (10)$$

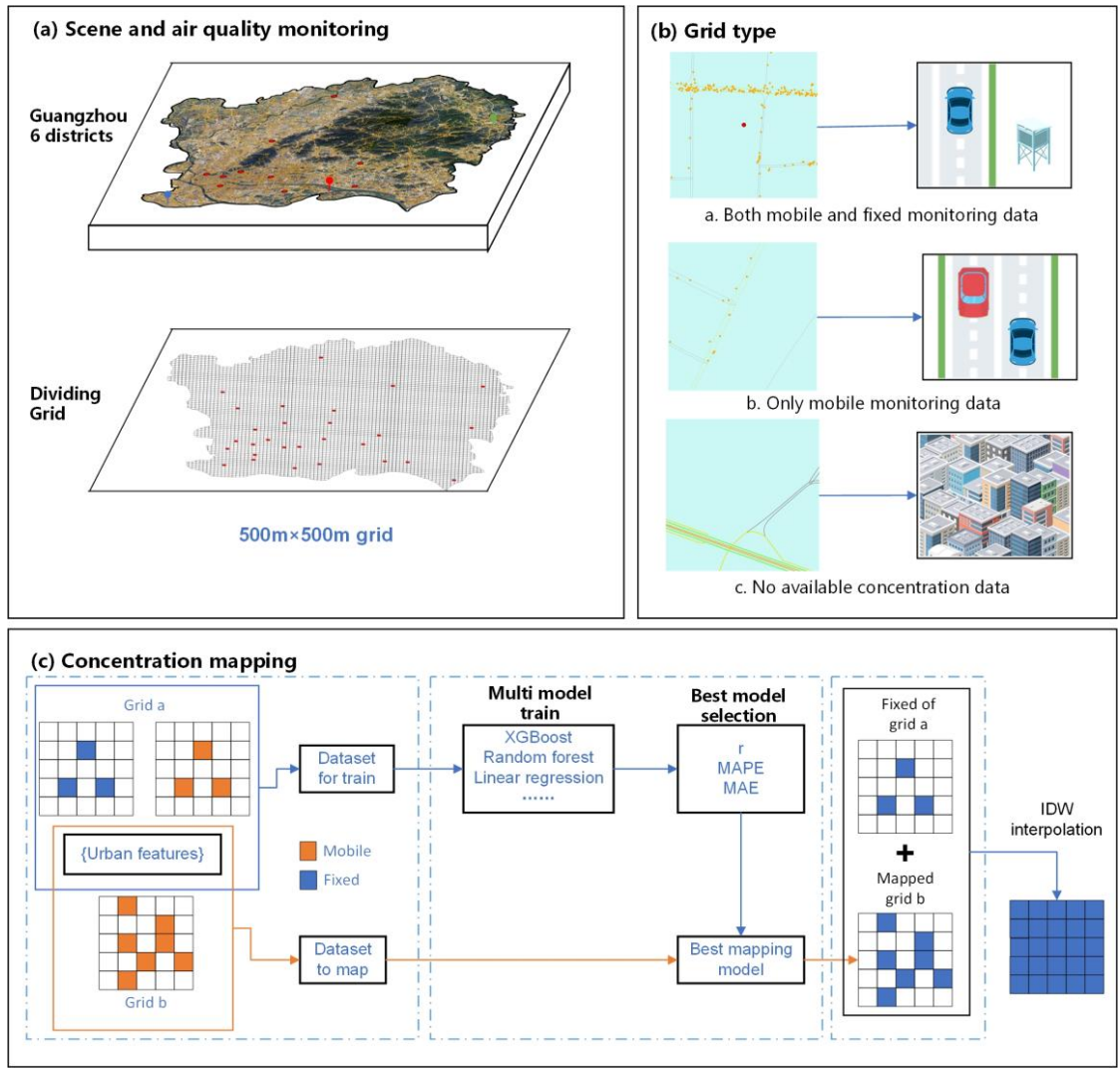

Fig.4 Construction of PM$_{2.5}$ pollution maps integrating mobile and fixed monitoring data.

In the final stage, three types of air pollution maps were constructed based on different monitoring data sources: (a) spatial interpolation only from fixed monitoring concentration data, (b) spatial interpolation only from mobile monitoring concentration data, and (c) spatial interpolation from mapped and fixed concentration data. The geographical interpolation method was inverse distance weighting (IDW).

# 3. Results

## 3.1. Calibration evaluation

The correlation analysis (Table.1) demonstrated that the r among the three LCSs exceeded 0.87, while the r between the AQMS and LCS was greater than 0.83. These values demonstrated a robust degree of concordance between the data obtained from fixed monitoring and that recorded by mobile LCS monitoring devices.

Table.1 The Pearson's correlation coefficients for LCS and AQMS 5-minute monitoring concentrations in the same position.

|  | AQMS | LCS 1 | LCS 2 | LCS 3 |
|---|---|---|---|---|
| **AQMS** | 1 | 0.86 | 0.84 | 0.83 |
| **LCS 1** | 0.86 | 1 | 0.89 | 0.87 |
| **LCS 2** | 0.84 | 0.89 | 1 | 0.99 |
| **LCS 3** | 0.83 | 0.87 | 0.99 | 1 |

Subsequently, AQMS data served as the standard reference for calibrating LCS

measurements. The adjusted results are displayed in Fig. 5 and Table. 2. Fig. 5(b) compared with (c), the results improved slightly when temperature was taken into account, which is consistent with previous studies showing that humidity was the main factor[42,43]. The results of XGBoost further show that the relationship between concentration, temperature and humidity is not simple linear, there is also a nonlinear deviation part.

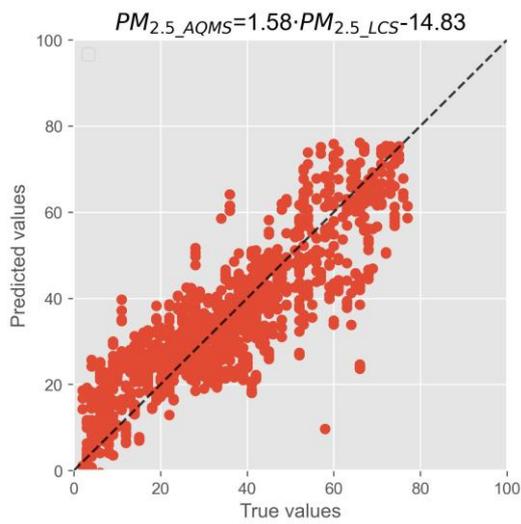

(a)

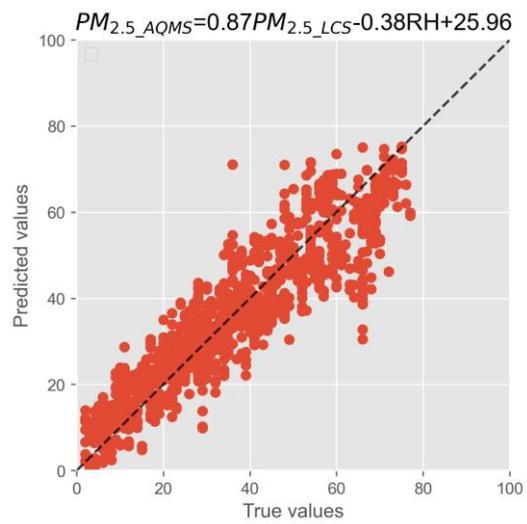

(b)

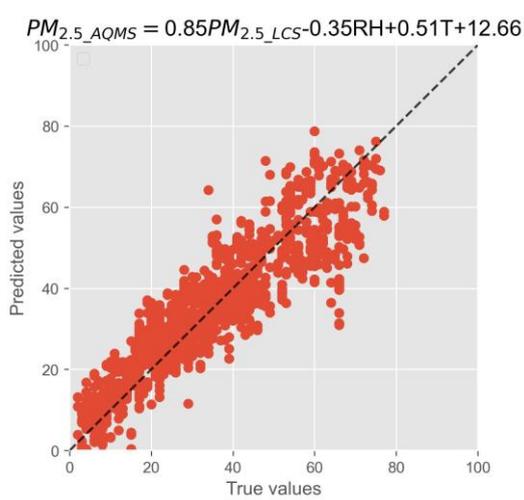

(c)

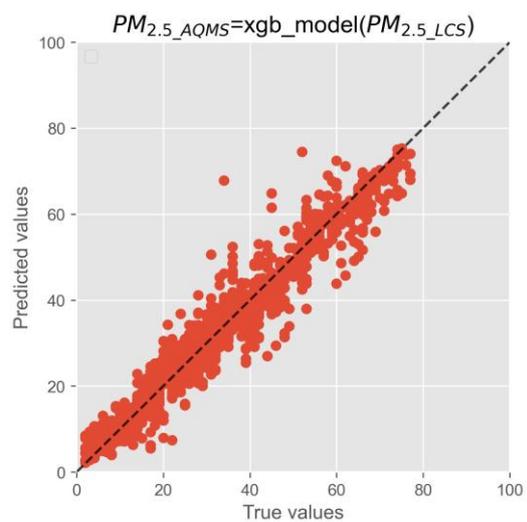

(d)

Fig. 5 Concentration correction results for LCS.

Table. 2 Metrics after calibration.

|     | r | $R^2$ | $MAE(\mu g/m^3)$ | $RMSE(\mu g/m^3)$ |
| --- | --- | --- | --- | --- |
| (a) | 0.85 | 0.63 | 7.09 | 9.42 |
| (b) | 0.9 | 0.78 | 5.88 | 7.63 |
| (c) | 0.91 | 0.78 | 5.84 | 7.54 |
| (d) | 0.97 | 0.94 | 2.95 | 4.25 |

## 3.2. Mapping model evaluation

A comparative analysis was conducted between the 52 fixed AQMS monitoring datasets and mobile LCS measurements over varying spatial proximities and temporal intervals to identify the most suitable spatiotemporal resolution for pollution mapping.

### 3.2.1. Model metrics

**Spatial-temporal correlation analysis:** The results are detailed in Table.3. In the case of one-hour resolution, it was observed that the r increased from 0.62 to 0.69 as the adjacent distance decreased from 2000 meter to 500 meter. Conversely, in the case of 5-minute resolution, the r exhibited a range of 0.63 to 0.71, indicating a more significant improvement compared to the one-hour resolution. Notably, the correlation between fixed and mobile monitoring was strongest when the adjacent distance was constrained to 500 meters. These findings support the selection of a 500-meter spatial resolution and a 5-minute temporal resolution, as they provide the highest level of spatial and temporal precision while maintaining an acceptable correlation coefficient.

Table.3 The Pearson correlation coefficient of fixed and mobile monitoring at different spatial and temporal neighbors.

| Distance | 5min | 10min | 30min | 60min |
|---|---|---|---|---|
| **500m** | 0.71 | 0.69 | 0.68 | 0.69 |
| **1000m** | 0.68 | 0.67 | 0.68 | 0.70 |
| **2000m** | 0.63 | 0.60 | 0.60 | 0.62 |

**Predicting model performance**: The performance of different machine learning-based mapping models in predicting $PM_{2.5}$ concentrations from mobile LCS data is summarized in Table.4. Results indicate that all trained models outperformed the baseline approach, which relied on averaged mobile monitoring concentrations for mapping. Among the evaluated models, the XGBoost-based mapping model exhibited the highest predictive performance, achieving r=0.822, MAE=6.474μg/m³, and MAPE=0.161. Compared to the baseline method of using averaged mobile monitoring concentrations, this model improved r improves by 7.6%, increasing from 0.764 to 0.822. The MAE decreases by 56.5%, dropping from 14.876μg/m³ to 6.474μg/m³, and the MAPE decreases by 57.5%, going from 0.379 to 0.161. While the increase in r was moderate, the reductions in MAE and MAPE were more substantial, highlighting the model's effectiveness in minimizing prediction errors. These results highlight the effectiveness of the machine learning approach in modeling the relationship between mobile and fixed monitoring concentrations within the local region.

Table.4 Performance of different mapping models in predicting $PM_{2.5}$ concentration.

| Model | MAE (μg/m3) | MAPE (%) | r |
|---|---|---|---|
| XGBoost | 6.474 | 0.161 | 0.822 |
| Random forest | 6.735 | 0.169 | 0.813 |
| Linear regression | 7.024 | 0.181 | 0.794 |
| Lasso | 7.077 | 0.183 | 0.790 |
| K neighbors nearest | 7.166 | 0.181 | 0.789 |
| Average | 14.876 | 0.379 | 0.764 |

### 3.2.2. Gain of urban features

Gain is an indicator to evaluate the influence of input parameters on the predict results of the mapping model, which can explain the importance of features. A higher gain value indicates a greater contribution of the corresponding feature to the overall information gain. This implies that the feature has a more substantial impact on enhancing the model's prediction performance. The results of XGBoost model for gains in input features were analyzed here. Among all the input features, the top five based on gain are as follows: average concentration (1341), primary road length (1191), tertiary road length (587), traffic area (487) and secondary road length (463). Therefore, the average concentration of mobile monitoring is a crucial factor in predicting the $PM_{2.5}$ concentration. However, it is important to note that different road lengths can greatly influence this mapping relationship. Additionally, other urban characteristic data also provide certain information gain, such as residential area (392), grass area (209) and building area (240).

## 3.3. Characteristics of PM$_{2.5}$ pollution

### 3.3.1. Hourly pollution

The variation and distribution patterns of PM$_{2.5}$ concentrations maps were analyzed using different temporal resolutions (hourly and 5-minute). March 1 was characterized by stable weather conditions, low wind speeds, and mild pollution levels. During the typical morning peak hour (9:00~10:00), the region exhibited an overall upward trend in PM$_{2.5}$ concentrations. Consequently, the selection of March 1, 7:00-19:00 for hourly changes, and specifically from 9:00-10:00 for 5-minute interval changes to facilitate the study.

The origin monitoring data of March 1 were aggregated into hourly and 5-minute intervals. The mobile monitoring data was mapped using the best-performing XGBoost model to predict the concentration of grids, referred to as the mapped concentration. Fig.6 and Fig. 8 illustrates the PM$_{2.5}$ concentration level of the raw monitoring data, including the hourly variation from 7:00-18:00 on the 1st and the 5-minute variation from 9:00-10:00 on the 1st.

In the selected hourly timeframe (Fig.6), the average concentration recorded by fixed AQMSs was 50.90±6.21μg/m$^3$, while mobile LCSs recorded an average of 67.60±7.42μg/m$^3$. This indicates a 32.78% increase in the average concentration measured by mobile LCSs compared to fixed AQMSs. Similarly, the mapped concentration was observed to be 51.08±2.88μg/m$^3$, which is closer to the fixed monitoring concentration. Overall, the mean concentrations obtained from mobile monitoring were consistently higher than those from fixed monitoring and mapped by XGBoost model.

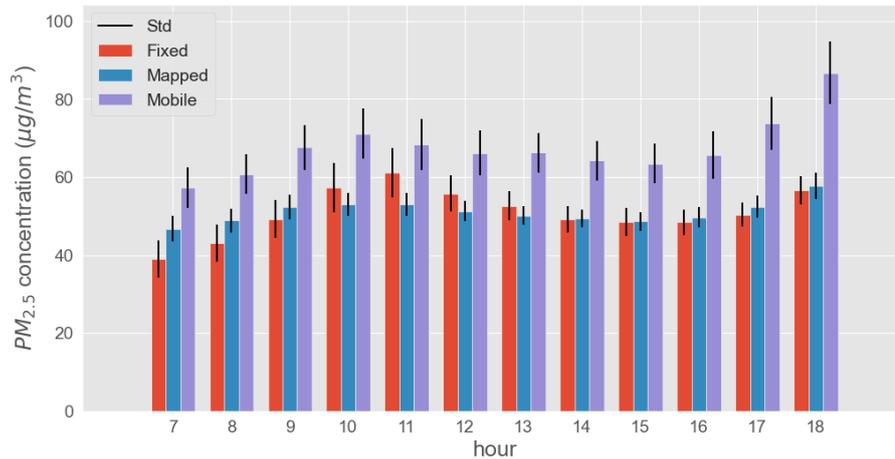

Fig.6 Hourly variation of the three concentration results at the selected time.

Fig.7 illustrates the spatial distribution in PM$_{2.5}$ concentrations from 7:00 to 18:00 on the March 1. Map (a) exhibits a clear centralized monitoring bias, while map (b) and (c) display a prominent speckle effect. The hourly variations in PM$_{2.5}$ concentrations exhibit consistent diurnal patterns, characterized by a systematic increase during specific intervals. Localized regions can also be observed to exhibit successive hourly variations, such as the decreasing (black square) of low concentration areas in the central region depicted in Fig.7. This phenomenon is particularly pronounced between 7:00 and 10:00, as illustrated in maps (a), (b), and (c), where a clear decline in PM$_{2.5}$ levels is observed.

The temporal variations in PM$_{2.5}$ concentrations were quantified as follows: Map

(a) showed a variation of 9.28±5.49%, map (b) exhibited 13.68±12.93%, and map (c) recorded 8.91±7.64%. Notably, the hourly maps derived from mobile monitoring data show significantly higher temporal and spatial fluctuations.

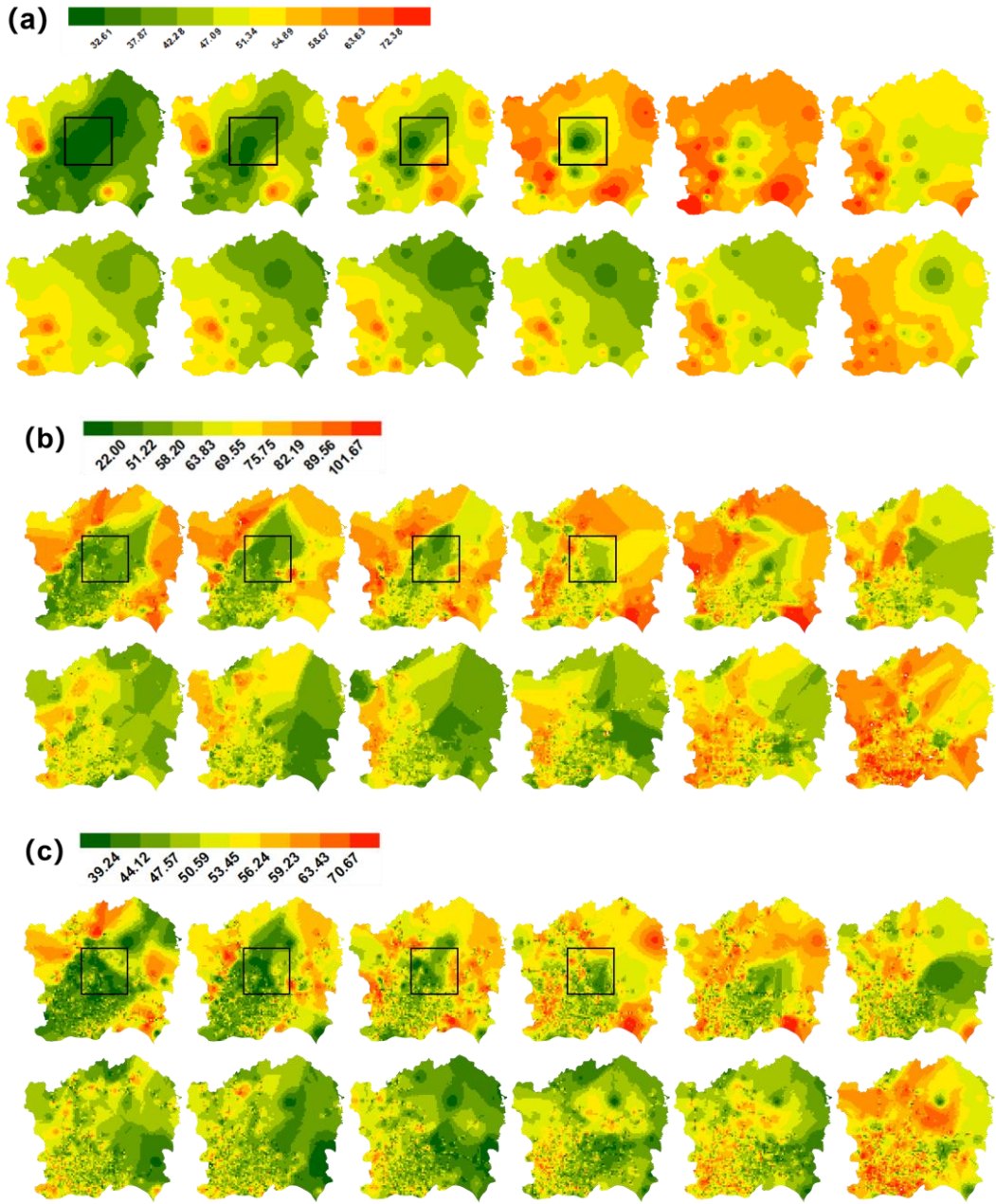

Fig.7 The hourly $PM_{2.5}$ concentration map from 7:00 to 18:00 on March 1, 2023. Unit: μg/m$^3$. (a) Fixed; (b) Mobile; (c) Mapped.

### 3.3.2. 5-minute pollution

During the selected 5-minute interval (Fig. 8), fixed monitoring data recorded an average concentration of 49.25±1.99μg/m$^3$, whereas the mobile monitoring reported 67.55±0.72μg/m$^3$, with mobile monitoring results 37.18% higher than those from fixed monitoring. The mapped PM$_{2.5}$ concentration, averaging 51.24±0.89μg/m$^3$, exhibits fluctuations that align more closely with the mobile monitoring concentration. Overall, PM$_{2.5}$ concentrations from mobile monitoring were found to be more than 30% higher than those from fixed across the different time scales. The mapped PM$_{2.5}$ concentration, however, is closer to the fixed monitoring compared to the mobile monitoring concentration. As the temporal resolution of observations was refined from hourly to 5-minute intervals, the variance in concentration fluctuations decreased significantly, allowing more subtle concentration changes to be detected at finer temporal scale.

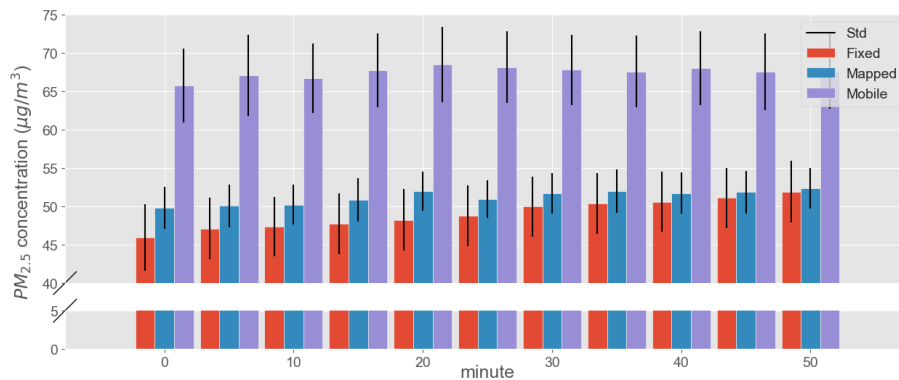

Fig. 8 The 5-minute variation of the three monitoring results at the selected time.

Fig.9 shows the variation results of the 5-minute maps. For fixed results (a), the entire concentration map varied considerably over the course of an hour, showing a steady upward trend. The average concentration computed from Fig.9(a) increased from 45.96μg/m$^3$ to 51.89μg/m$^3$, a rise of 12.90%. However, the adjacent 5-minute change was relatively small and insignificant in most areas, with an average adjacent mean absolute concentration percentage variation of 1.12±0.73%. The mobile maps (Fig.9b) had higher concentration level compared to the fixed, from 54.61μg/m$^3$ at 9:00, gradually increasing to 59.50μg/m$^3$ at 10:00, an overall increase of 8.95%. However, the highest concentration during that hour occurred at 9:40 at 61.85μg/m$^3$, up 13.26% compared to the initial. In contrast, the 5-minute average absolute concentration percentage changed for the mobile maps was 3.15±2.44%, with more pronounced fluctuations compared to the fixed. This is consistent with fluctuations in the raw monitoring data. The periphery of this concentration map, especially the upper right region, presented a single concentration value, and these areas were consistent with the areas in Fig.2 where mobile monitoring data were missing. The initial average concentration of the mapped maps (Fig.9c) was 49.82μg/m$^3$, which was 8.40% above the fixed maps and 8.77% below the mobile maps. The last mapped concentration was 51.90μg/m$^3$, which was 0.02% above the fixed maps and 12.77% below the mobile. The adjacent variation was 1.01±0.65%. The 5-minute concentration fluctuation results exhibited closer alignment with the fixed monitoring data and demonstrated greater stability compared to the mobile monitoring outputs. Notably, the periphery regions of the concentration map displayed subtle variations, contrasting with the more uniform patterns observed in Fig.9(b). There was no longer a single concentration value, and there appeared to be a gentle fluctuation in concentration. These findings suggest that the refined temporal resolution not only enhances the stability of concentration measurements but also reveals finer spatial gradients in air pollution, which are often masked in coarser datasets.

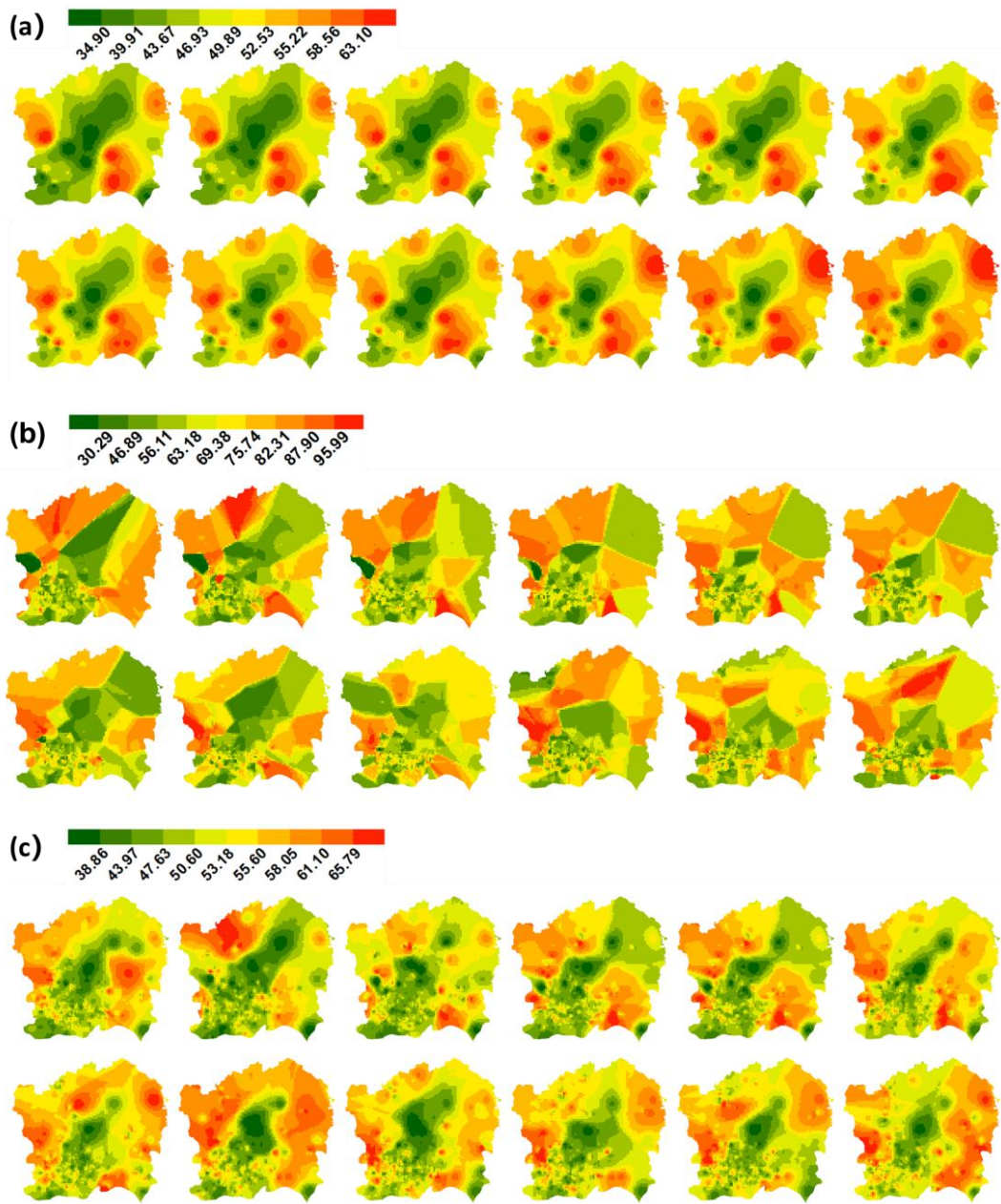

Fig.9 The 5-minute PM$_{2.5}$ concentration map from 9:00 to 10:00 on March 1, 2023. Unit: μg/m$^3$.

(a) Fixed; (b) Mobile; (c) Mapped.

### 3.3.3. Map characteristics from different monitoring data

Consistent temporal trends in $PM_{2.5}$ concentrations are evident across the datasets derived from different monitoring methodologies. As illustrated in Fig.7 and Fig.9, the concentration depicted in map (b) (mobile monitoring) is significantly elevated compared to the other two maps. The map results indicate that, across different scales, the concentration levels observed through mobile monitoring are consistently higher than those obtained through the fixed monitoring, a finding that aligns with the original data from both monitoring approaches.

On a temporal scale, the rate of concentration variation was notably reduced in the 5-minute resolution maps compared to the hourly maps. The higher temporal resolution of the 5-minute data facilitated a more continuous and granular representation of $PM_{2.5}$ fluctuations over time. Map(a) exhibits the most pronounced improvement in time continuity, with a transition from 9.28±5.49% to 1.12±0.73%.

Clear spatial pattern can be found evidently in all map results, with a global Moran's index exceeding 0.72. The map(c) is expected to exhibit greater spatial discreteness compared to maps(a) and (b). The 5-minute maps, however, are anticipated to exhibit slightly higher volatility and slightly lower spatial discretization. The 5-minute maps demonstrate an improvement in their level of spatial autocorrelation, with the Moran's index for map (b) increasing from 0.89 to 0.94, and for map (c), rising from 0.79 at the hourly level to 0.90.

# 4. Discussions

A substantial volume of $PM_{2.5}$ concentration data was collected using mobile LCS and fixed AQMS. By analyzing the spatial and temporal distribution of the collected data, an optimal resolution of 5-minute and 500-meter was identified. Utilizing urban feature data, machine learning models were trained to infer the mapping relationship between mobile and fixed monitoring concentrations. Finally, $PM_{2.5}$ concentration maps with varying spatial-temporal characteristics were generated using different monitoring data sources.

Mobile monitoring from LCS provides the majority of raw concentration data for constructing the pollution map. The incorporation of mobile monitoring data from a wider range of spatial locations has alleviated the centralized bias and the smoothing effect, which were previously induced by the sparsity of fixed monitoring sites. However, the map results are also influenced by the characteristics of mobile monitoring data (Fig. 10). Urban mobile monitoring tends to concentrate data collection within core areas, while peripheral regions experience significantly lower coverage. As monitoring extends outward from the urban center, the prevalence of unmonitored zones increases, leading to spatial disparities in data availability. Other Chinese megacities such as Beijing, Shanghai, and Chengdu also show the same spatial characteristics of large-scale mobile monitoring data [31,44,45]. Mobile monitoring data acquisition is highly dependent on road network distribution, with significant limitations in areas devoid of road infrastructure. The availability of roads serves as a key determinant of whether a specific location can be captured. To optimize the spatial coverage of mobile monitoring with a limited number of devices, it is crucial to select taxis with varied route patterns. An assessment of urban features as influencing factors indicated that roads exerted the most significant impact, followed by buildings and green spaces.

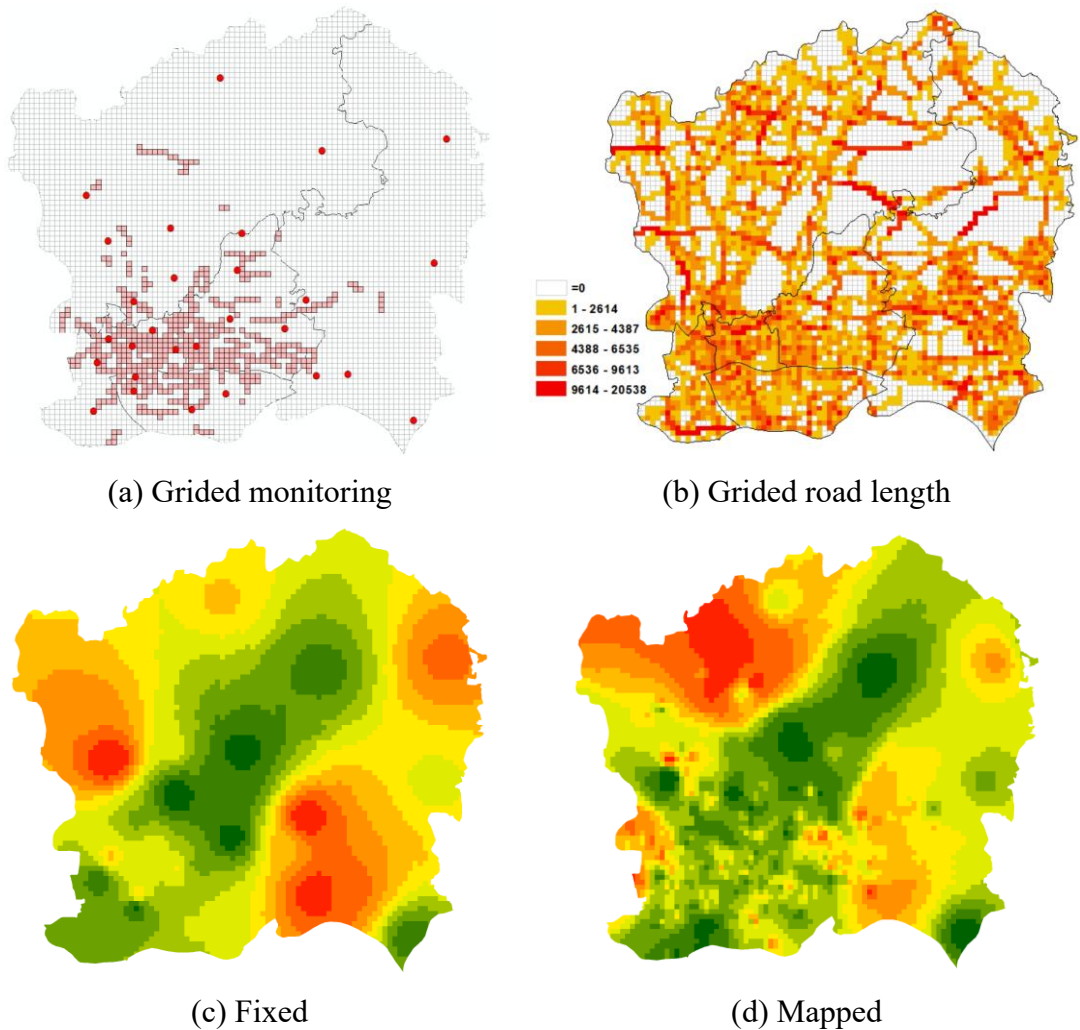

(a) Grided monitoring  (b) Grided road length

(c) Fixed  (d) Mapped

Fig. 10 Grided monitoring and road distribution and map results at 9:00, March 1.

Finer temporal resolutions are feasible when monitoring systems generate adequate data to support detailed analysis of pollution evolution. Although fixed monitoring can mitigate spatial gaps in mobile monitoring, concentration data in some peripheral urban areas remain limited. Therefore, urban managers can make up for the lack of pollution awareness by increasing fixed monitoring in areas with sparse mobile monitoring. A thorough understanding of the characteristics of each monitoring data type and their interrelationship is essential for effectively integrating mobile and fixed monitoring data, thereby enhancing methods for constructing urban air pollution maps. The mapped concentration, however, closely resemble those based on fixed monitoring concentrations, which can be attributed to the use of fixed monitoring concentrations as targets in the mapping process.

Concerns should be raised regarding short-term and hyperlocal high pollution levels, as exposure to such levels can still have a significant impact on the human health [46,47]. Enhanced spatiotemporal resolution improves the capture of short-term variations in air pollution, facilitating timely responses to sudden pollution events and enabling residents to promptly evaluate air quality conditions and adopt necessary protective measures. The distribution of local spatial pollution varies greatly in city level [48], high-resolution pollution concentration modeling is essential for capturing fine-scale variations in air quality. Combined with the population activity trajectory data, the individual refined exposure risk assessment research can be further carried out.

Additionally, the 5-minute concentration maps provide improved continuity and completeness, aiding in the identification and analysis of trends and cyclical changes over extended periods.

Despite its contributions, this study has several limitations. Mobile LCS monitoring is characterized by significant temporal and spatial variability, resulting in lower data coverage in non-hotspot urban areas. This issue is particularly evident in peripheral regions, where the lower density of the road network limits the volume of data collected by taxis. Regions lacking road infrastructure, including bodies of water and forests, have no monitoring data, resulting in reduced representativeness and spatial gaps in the dataset for these peripheral areas. Additionally, the current study relies on the stability of fixed and mobile monitoring methods. Future research could explore the use of neural network techniques to better integrate and optimize the data from mobile and fixed monitoring, potentially yielding more accurate predictive results. In the process of concentration mapping, dynamic factors such as wind direction, and rainfall can be further considered.

## 5. Conclusions

Machine learning enables the integration of monitoring concentrations from mobile LCS and fixed AQMS by mapping their relationship. This eliminates the centralized monitoring bias caused by the excessively sparse fixed monitoring points, and also corrects the concentration offset of the taxi-based mobile monitoring. A time scale of 5-minute exhibits greater continuity and stability compared to hourly. The acquisition of a greater quantity of monitoring data allows for the capture of refined spatial variations in pollution that are closely associated with the distribution of the traffic road network. These results provide significant insights into the effectiveness of large-scale LCS mobile monitoring for pollution mapping and highlight potential avenues for future research. Furthermore, they enhance the precision of urban pollution assessment and support strategies for promoting green and sustainable urban development.

# CRediT authorship contribution

**Rui Xu:** Writing – original draft & review & editing, Methodology, Data curation, Conceptualization. **Dawen Yao:** Methodology, Investigation. **Yuzhuang Pian.**: Writing – original draft, Investigation. **Ruhui Cao:** Writing – original draft, Methodology. **Yixin Fu:** Methodology, Conceptualization. **Xinru Yang:** Data curation. **Ting Gan:** Data curation. **Yonghong Liu:** Writing – review & editing, Funding acquisition, Supervision, Resources, Conceptualization.

of Things, Procedia Computer Science 157 (2019) 638–645. https://doi.org/10.1016/j.procs.2019.08.224.

[22] K.H. Kim, K.-H. Kwak, J.Y. Lee, S.H. Woo, J.B. Kim, S.-B. Lee, S.H. Ryu, C.H. Kim, G.-N. Bae, I. Oh, Spatial Mapping of a Highly Non-Uniform Distribution of Particle-Bound PAH in a Densely Populated Urban Area, Atmosphere 11 (2020) 496. https://doi.org/10.3390/ATMOS11050496.

[23] C. Tong, Z. Shi, W. Shi, P. Zhao, A. Zhang, Mapping Microscale PM2.5 Distribution on Walkable Roads in a High-Density City, IEEE J. Sel. Top. Appl. Earth Observations Remote Sensing 14 (2021) 6855–6870. https://doi.org/10.1109/JSTARS.2021.3075442.

[24] J. Fu, D. Tang, M.L. Grieneisen, F. Yang, J. Yang, G. Wu, C. Wang, Y. Zhan, A machine learning-based approach for fusing measurements from standard sites, low-cost sensors, and satellite retrievals: Application to NO2 pollution hotspot identification, Atmospheric Environment 302 (2023) 119756. https://doi.org/10.1016/j.atmosenv.2023.119756.

[25] S. Wang, Y. Ma, Z. Wang, L. Wang, X. Chi, A. Ding, M. Yao, Y. Li, Q. Li, M. Wu, L. Zhang, Y. Xiao, Y. Zhang, Mobile monitoring of urban air quality at high spatial resolution by low-cost sensors: impacts of COVID-19 pandemic lockdown, Atmos. Chem. Phys. 21 (2021) 7199–7215. https://doi.org/10.5194/acp-21-7199-2021.

[26] L. Morawska, P.K. Thai, X. Liu, A. Asumadu-Sakyi, G. Ayoko, A. Bartonova, A. Bedini, F. Chai, B. Christensen, M. Dunbabin, J. Gao, G.S.W. Hagler, R. Jayaratne, P. Kumar, A.K.H. Lau, P.K.K. Louie, M. Mazaheri, Z. Ning, N. Motta, B. Mullins, M.M. Rahman, Z. Ristovski, M. Shafiei, D. Tjondronegoro, D. Westerdahl, R. Williams, Applications of low-cost sensing technologies for air quality monitoring and exposure assessment: How far have they gone?, Environment International 116 (2018) 286–299. https://doi.org/10.1016/j.envint.2018.04.018.35